\documentclass[journal]{IEEEtran}

\ifCLASSINFOpdf
\else
   \usepackage[dvips]{graphicx}
\fi

\usepackage{url}
\usepackage{graphicx}
\usepackage{xcolor}
\usepackage{cite}
\usepackage[colorlinks=true]{hyperref}
\usepackage{amsmath}
\usepackage{amssymb}
\usepackage{booktabs}
\usepackage{comment}
\usepackage{multirow}
\usepackage{amsbsy}
\usepackage{subcaption}

\newcommand{\tit}[1]{\smallbreak\noindent\textbf{#1.}}

\def \ie {\emph{i.e. }}
\def \eg {\emph{e.g. }}
\def \etal {\emph{et al.}}

\begin{document}

\title{Unveiling the Truth:\\Exploring Human Gaze Patterns in Fake Images}

\author{Giuseppe Cartella, Vittorio Cuculo, Marcella Cornia, Rita Cucchiara
\thanks{This work was supported by the PNRR project Italian Strengthening of Esfri RI Resilience (ITSERR) funded by the European Union - NextGenerationEU (CUP: B53C22001770006).}
\thanks{G. Cartella, V. Cuculo, and R. Cucchiara are with the Department of Engineering ``Enzo Ferrari'', University of Modena and Reggio Emilia, Modena, Italy (e-mail: \{giuseppe.cartella, vittorio.cuculo, rita.cucchiara\}@unimore.it).}
\thanks{M. Cornia is with the Department of Education and Humanities, University of Modena and Reggio Emilia, Reggio Emilia, Italy (e-mail: marcella.cornia@unimore.it).}}

\markboth{IEEE Signal Processing Letters}
{Cartella \MakeLowercase{\textit{et al.}}: Unveiling the Truth: Exploring Human Gaze Patterns in Fake Images}
\maketitle

\begin{abstract}
Creating high-quality and realistic images is now possible thanks to the impressive advancements in image generation. A description in natural language of your desired output is all you need to obtain breathtaking results. However, as the use of generative models grows, so do concerns about the propagation of malicious content and misinformation. Consequently, the research community is actively working on the development of novel fake detection techniques, primarily focusing on low-level features and possible fingerprints left by generative models during the image generation process. In a different vein, in our work, we leverage human semantic knowledge to investigate the possibility of being included in frameworks of fake image detection. To achieve this, we collect a novel dataset of partially manipulated images using diffusion models and conduct an eye-tracking experiment to record the eye movements of different observers while viewing real and fake stimuli. A preliminary statistical analysis is conducted to explore the distinctive patterns in how humans perceive genuine and altered images. Statistical findings reveal that, when perceiving counterfeit samples, humans tend to focus on more confined regions of the image, in contrast to the more dispersed observational pattern observed when viewing genuine images. Our dataset is publicly available at: \url{https://github.com/aimagelab/unveiling-the-truth}.
\end{abstract}


\IEEEpeerreviewmaketitle

\section{Introduction}
\IEEEPARstart{O}{ne} of the most recent groundbreaking advancements in the realm of image generation concerns the advent of diffusion models~\cite{sohl2015deep,ho2020denoising,rombach2022high,song2020denoising,dhariwal2021diffusion,nichol2021improved} which have swiftly garnered significant acclaim within the scientific community, marking a new era in the field of generative artificial intelligence.
The impressive ability to generate high-quality and realistic content has continued to advance, and the adoption in various contexts including content creation~\cite{ruiz2023dreambooth,saharia2022photorealistic,ramesh2021zero,karras2020analyzing} and image enhancement~\cite{yi2023diff,guo2023shadowdiffusion} is growing at a steady pace. In addition, the training of increasingly large deep networks can be empowered by the availability of a massive volume of synthetic data.

\begin{figure}[t!]
    \centering
    \includegraphics[width=\linewidth]{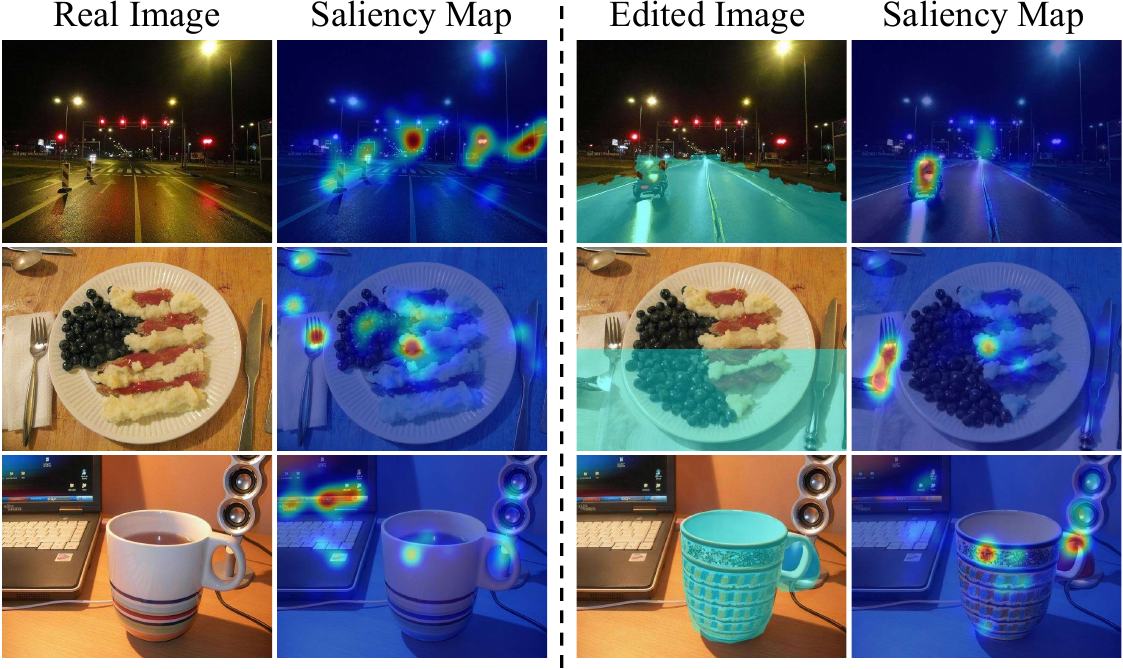}
    \caption{Overview of the human gaze patterns when observing real and altered images. 
    Interestingly, humans tend to focus on more circumscribed areas when looking at counterfeit samples. Light-blue masks of edited images represent inpainted regions.}
    \label{fig:teaser}
    \vspace{-.35cm}
\end{figure}

However, the proliferation of false and malicious content poses serious challenges and highlights the importance of distinguishing genuine content from synthetic ones. With this aim, researchers are putting active efforts into developing novel fake detection techniques~\cite{ojha2023towards,corvi2023intriguing,wang2020cnn,wu2023generalizable,jeong2022frepgan}.
Most of the computer vision literature focuses on the recognition of fake images and videos, with a particular emphasis on face manipulation~\cite{rossler2019faceforensics++,zhao2021multi,li2018exposing}.  However, recent trends have emerged that extend the recognition to natural images (\ie~landscapes, urban scenes, etc.)~\cite{amoroso2023parents,lorenz2023detecting}, albeit still at an early stage.

Typically, state-of-the-art techniques ground the detection of fake samples on the analysis of the generative models' feature space. Various studies~\cite{frank2020leveraging,corvi2023detection} demonstrated that images stemming from different generative models present discernible fingerprints left behind by the model during the generation process.
Wang \etal~\cite{wang2020cnn} showed that by adopting a proper post-processing and data augmentation pipeline it is possible to generalize across different GAN-based models. However, the introduction of diffusion models shed light on a major challenge, highlighting how generalization across different families of generative models is still an open research problem~\cite{ojha2023towards,corvi2023detection}.
From a semantic perspective, as assisted in the field of face manipulation, modifications can be carried out starting from real images and altering only portions of them in order to obtain a result that preserves the original context, making fake detection an increasingly challenging task.

In this paper, drawing inspiration from visual attention literature~\cite{foulsham2010asymmetries,parkhurst2002modeling,peters2005components}, we bring humans into the loop with the aim to exploit their semantic knowledge and generalisation skills acquired through evolution and further developed in a lifelong learning process. In particular, we seek an answer to the following research question:
\begin{itemize}
    \item Does an underlying attentive pattern exist governing the human visual perception when looking at partially manipulated images compared to genuine ones?
\end{itemize}

We posit the existence of such a disparity and, to validate this assertion,
we begin by collecting authentic samples from various existing datasets. For each image under investigation, we produce three distinct altered variants through the implementation of as many editing methodologies based on state-of-the-art diffusion models~\cite{rombach2022high,brooks2023instructpix2pix}.
To assess human visual perception in the context of authentic samples versus their counterfeit equivalents, an eye-tracking experiment is further devised. During this experiment, a sequence of images is presented to the participants, and their ocular movements are recorded as they attempt to distinguish real from fake samples.

Finally, a statistical analysis of the collected fixations is conducted, which reveals significant disparities in the viewing patterns that occur when looking at genuine and fake images. In particular, when analyzing the entropy distribution of the acquired saliency maps, we find that fake images elicit high fixation concentrations in specific regions, resulting in lower entropy values when compared to their original counterpart (Fig.~\ref{fig:teaser}).
As a result, we draw the conclusion that humans tend to direct their attention on more delimited areas when perceiving manipulated images.
We believe our preliminary results open up further research towards the integration of human gaze information within automatic fake detection pipelines.

\section{Proposed Approach}
Given a set of real and fake images, our goal is to conduct a statistical analysis to investigate the existence of an underlying pattern governing the visual perception of partially manipulated images, enabling further studies on how the human gaze could improve the existing fake detection models. 
With this aim, and due to the lack of an existing dataset in the literature, we collect images from different sources encompassing a variety of scenes ranging from indoor to outdoor environments. 
\subsection{Dataset Collection}
We define a set of stimuli that cover scenes and environments of varying degrees of complexity. In our dataset, three distinct categories can be identified: \textit{indoor}, \textit{outdoor urban}, and \textit{outdoor natural}.
The first two categories usually exhibit a plethora of intricate details, a wide set of objects, and a rich diversity of color contrasts. On the opposite, the outdoor natural category features a lower amount of salient objects, scarce occurrences of high-frequency details, and a prevalence of more uniform color palettes.
Such diversity strongly influences the way people perceive images \cite{marr1976analyzing,oliva2006building}, thus guaranteeing appropriate data heterogeneity.
Images are extracted from three publicly available datasets, namely COCO~\cite{lin2014microsoft}, ADE20K~\cite{zhou2017scene}, and LHQ~\cite{skorokhodov2021aligning}.
Since text, faces, and animals are known to be very salient~\cite{cerf2007predicting,judd2009learning,cerf2009faces}, we filter out all the images including these three classes to avoid any possible bias, following recent literature~\cite{yang2020predicting}.
In the filtering process, we also discard low-resolution images for all datasets, keeping those with a minimum size of $640\times480$ (or $480\times640$).

\subsection{Image Editing Techniques}

\begin{figure*}[t!]
    \centering
    \begin{subfigure}[t]{0.33\textwidth}
        \centering
       \includegraphics[height=0.7\textwidth]{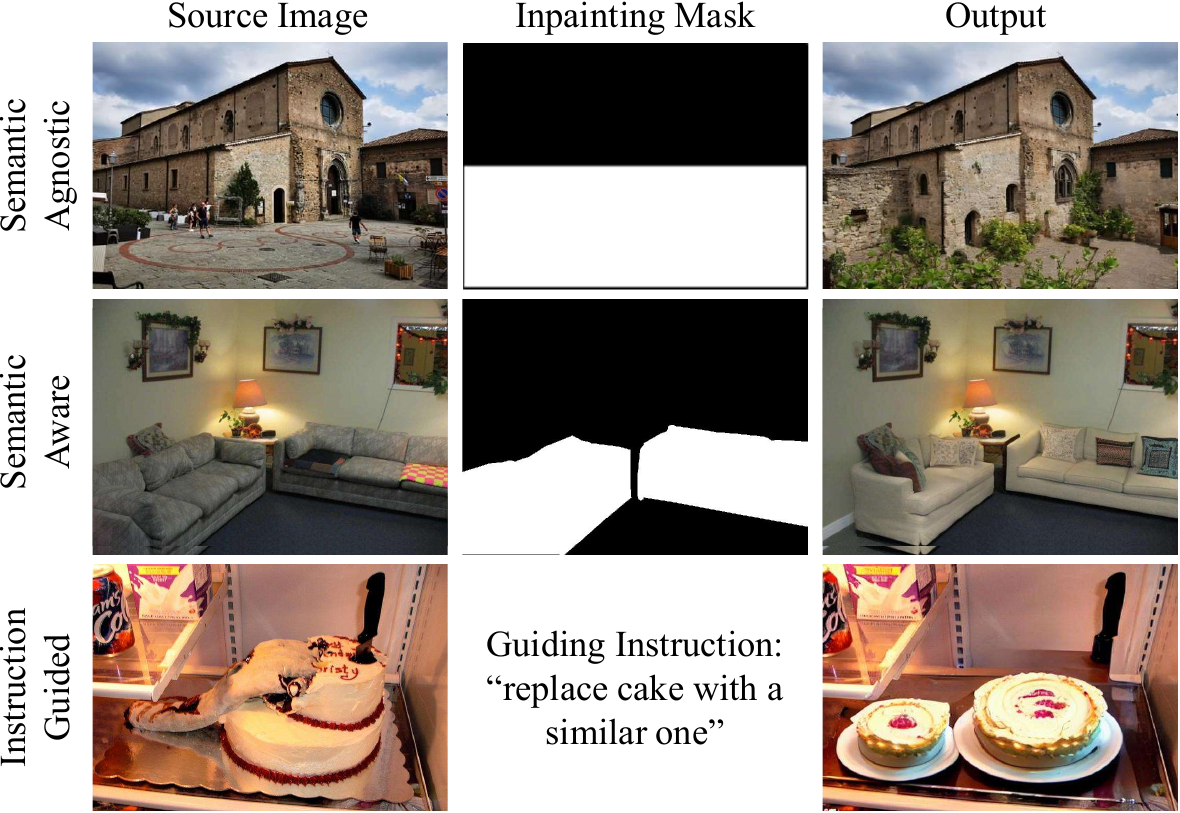}
        \caption{}
        \label{fig:editings}
    \end{subfigure}
    \hfill
    \begin{subfigure}[t]{0.31\textwidth}
        \centering
       \includegraphics[height=0.708\textwidth]{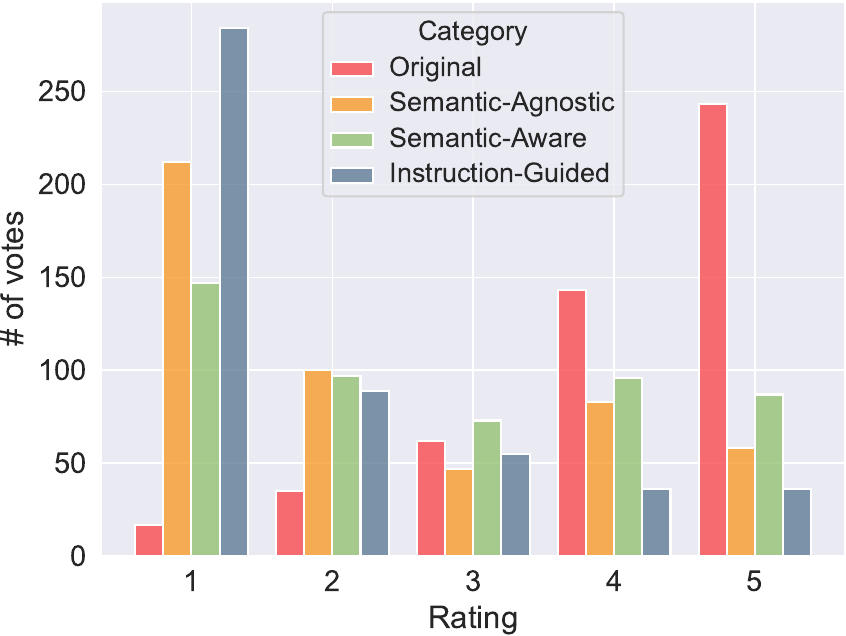}
        \caption{}
        \label{fig:ratings}
    \end{subfigure}
    \hfill
    \begin{subfigure}[t]{0.31\textwidth}
        \centering
       \includegraphics[height=0.708\textwidth]{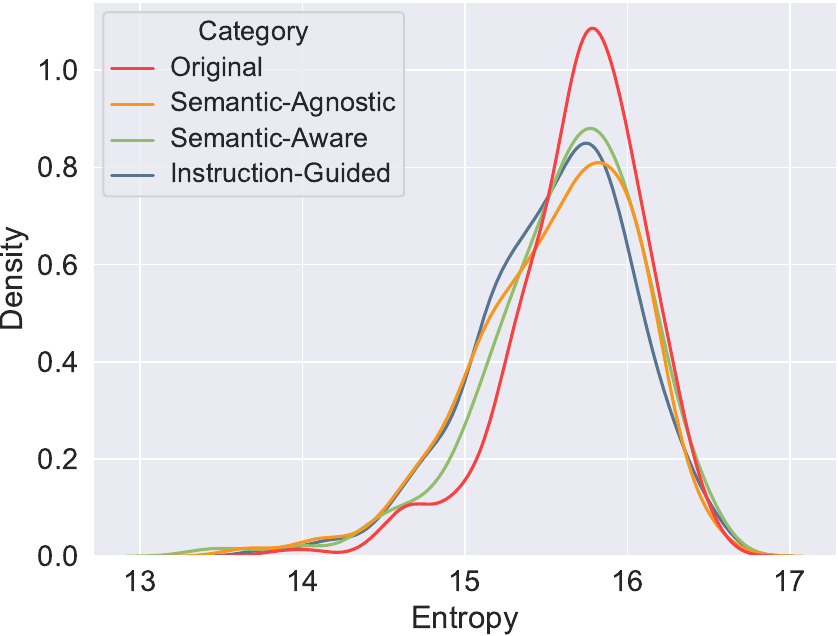}
        \caption{}
        \label{fig:entropy}
    \end{subfigure}
    \vspace{-.1cm}
       \caption{Qualitative visualizations of the proposed approach. (a) Image editing examples where the white masks represent the inpainting regions. (b) Histogram of the ratings of realism given by the users in the eye-tracking experiment. (c) Kernel density estimation of the saliency maps' entropy across viewers.}
       \label{fig:three graphs}
        \vspace{-.35cm}
\end{figure*}

In contrast to prior literature on deepfake detection which
has been focusing on recognising entirely generated images~\cite{ojha2023towards,corvi2023detection,amoroso2023parents}, we concentrate on manipulating images in a subtle manner, preserving the realism, semantics, and context of the original image, thus making the deepfake detection task even more challenging. 
To this end, three different types of intervention are implemented (see Fig.~\ref{fig:editings}).

\tit{Semantic-Agnostic (SA)}
Given a real source image $I$, we desire to create its fake counterpart $\Tilde{I}$. This task can be formally classified as an inpainting problem where the inpainting region is defined as a binary mask $M$, constructed by randomly masking half of the image $I$. Inspired by~\cite{aboutalebi2023deepfakeart}, for each sample, we randomly inpaint one among the following parts: bottom, upper, left, right, upper left diagonal, upper right diagonal, bottom left diagonal, bottom right diagonal, or random patches (until we cover at least $50\%$ of the image).
As an inpainting model, we adopt the Stable Diffusion v2.0 inpainting pipeline\footnote{\url{https://huggingface.co/stabilityai/stable-diffusion-2-inpainting}}~\cite{rombach2022high} which we refer to as $f_\theta$. To enhance the final output and avoid the generation of undesired objects, we make use of negative prompts $N$.
Notably, in our experiments, we qualitatively find that having no guiding caption $c$ as input leads to more realistic outputs.
Formally, the complete inpainting process can be defined as
$\Tilde{I} = f_\theta(I, M, c=\emptyset, N)$.

\tit{Semantic-Aware (SW)} Although semantic-agnostic manipulation leads to realistic results, there is no control over the generated output. 
As a different image editing technique, we propose semantic-aware inpainting, where objects present in the scene are substituted with others of the same class (\eg replace a bed with another bed).
This choice is driven by the intent to preserve the context and semantics of the scene, and ensure that the generated object well fits the given inpainting mask.
With respect to the previous manipulation category, edits affect smaller areas and are more intricate to discern.
To define the inpainting region, we start from the image segmentation maps of the images. COCO and ADE20K already provide a segmentation mask for each object $O_i$ and the corresponding textual class label $y$, while for LHQ, we construct the needed information by making use of the RAM-Grounded-SAM model\footnote{\url{https://github.com/IDEA-Research/Grounded-Segment-Anything}}~\cite{kirillov2023segment,zhang2023recognize}. 
We designate as an inpainting region of the real image $I$, the binary mask $M_{O^*}$ corresponding to a randomly chosen object $O^*$ among those with an occupancy between $10\%$ and $40\%$ of the total image area. The generation process is performed in the same manner as the \textit{semantic-agnostic}, but to guarantee context preservation, the guiding caption $c$ for the Stable Diffusion model is the textual class label $y$ of the selected object. Negative prompts $N$ are maintained the same. Overall, we obtain
$\Tilde{I} = f_\theta(I, M_{O^*}, c=y, N)$.

\tit{Instruction-Guided (IG)}
Recently, the field of generative artificial intelligence has witnessed the emergence of innovative generative methods, with instruction-guided image editing standing out as a prominent current research direction~\cite{zhang2023hive,brooks2023instructpix2pix}.
In our context, we apply the InstructPix2Pix model\footnote{\url{https://huggingface.co/vinesmsuic/magicbrush-jul7}}~\cite{brooks2023instructpix2pix}, able to follow a given editing instruction in natural language to produce the manipulated version of the given sample. 
In our problem, we feed InstructPix2Pix, referred to as $g_\theta$, with the source input $I$, and the following editing instruction $c$: \texttt{replace \{y\} with a similar one}. Specifically, \texttt{y} represents the textual class label of an object selected by following the same procedure of the \textit{semantic-aware} editing technique.
In our implementation, we adopt a finetuned version of InstructPix2Pix~\cite{zhang2023magicbrush}, which has been shown to produce better images according to human evaluation. The final synthesised image
is defined as $\Tilde{I} = g_\theta(I, c(y))$.
\subsection{Eye-Tracking Experiment}
Acquiring eye-tracking data is crucial to understand
the patterns governing the way humans look at real and fake stimuli. With this aim, we set up an eye-tracking experiment involving 20 participants.
Each person sits in front of a screen with a resolution of $1920\times1080$ pixels, equipped with a screen-based eye tracker, at a distance of $68$ cm.
The screen size is $54$ cm $\times$ $30.3$ cm.
To accommodate the inherent variability in the perceptual process of different people, we ensure that the image is viewed by five different observers.
Each stimulus is shown for 5 seconds, 
and the user is instructed to carefully observe the presented image while assessing its authenticity. Following this time-lapse, the stimulus is replaced with a rating screen where the user is 
required to evaluate the realism by selecting a numerical value on a 5-point Likert scale.
The lower end denotes a greater level of confidence in the image's artificial nature, whereas the upper range suggests a strong belief in the image's authenticity.
Prior to presenting the next stimulus, a grey screen featuring a small black cross at the center is exhibited for one second, aimed at engaging the user's attention.
To ensure high-quality acquisition, we calibrate the eye-tracker at the beginning of each experiment. 

Overall, the experiment is based on a total of 400 stimuli, comprising 100 unique genuine samples, uniformly distributed between the three datasets and their respective edited counterparts. For each observer, we randomly choose 100 images and, to avoid any bias, we ensure that at most one instance of the same image is shown (\emph{i.e.} if an observer is presented a semantic-agnostic edited image, then no other versions of the same image appear to the same participant).

\section{Experimental Analysis}
We conduct an in-depth statistical analysis of the recorded fixations to assess the evidence of a distinguishable pattern in the gaze behavior of individuals when viewing genuine versus counterfeit images. Overall, the analysis is performed on 2,000 recorded human scanpaths.

\tit{Human Annotations} 
In the first part of the investigation, we examine how users' ratings are distributed across the images.
Fig.~\ref{fig:ratings} depicts the histogram of the users' realism perception. 
In general, the observers are able to distinguish 
the real nature of most of the images.
A considerable number of fake images are given a low rating (\ie~1 or 2), and the majority of genuine samples are correctly identified. 
Nonetheless, some real images are classified as altered or possibly altered (\ie~ratings 1, 2, or 3). Such results can be ascribed to the bias introduced by the requested task or the nature of the image itself (\eg~a badly taken picture or out-of-context objects).
On the same line, occurrences of improperly categorized edited images (\ie~ratings 4 or 5) are observable, meaning that the adopted generative models, in several cases, synthesize highly realistic content.
As a final consideration, we point out that the instruction-guided category is the easiest to detect for humans, while the opposite holds for semantic-aware editing, leading to the most realistic outputs.
The explanation behind such a result is that the generation through InstructPix2Pix is much more challenging. Indeed, reference inpainting masks are absent and our constructed textual prompt does not heavily constrain the generation process.
As a consequence, such methodology is more prone to the creation of artifacts or out-of-context details. On the contrary, semantic-aware editing represents a more constrained type of intervention, with limited inpainting areas and a textual caption that guides the generation of a specific object that surely fits the context by construction.

\tit{Saliency Entropy} Another important analysis is the one regarding the eye-gaze pattern, enabling the study of the perceptual response of individuals to stimuli of different natures. 
As a first step,
we adopt the velocity-threshold fixation identification algorithm proposed in~\cite{salvucci2000identifying} to distinguish between fixations and saccades. The algorithm computes point-to-point velocities and classifies each raw point as fixation or saccade based on a simple velocity threshold.

To evaluate the consistency of human fixations over an image, we measure the entropy of the saliency maps across all the observers.
Given a sample image and its recorded fixations, a ground truth saliency map is obtained by convolving a fovea-sized Gaussian kernel over all the fixations locations.

Empirical distributions of the saliency maps' entropy for each of the four considered categories are reported in Fig.~\ref{fig:entropy}.
In a qualitative assessment, a significant distinction is noticeable between the distributions of real and fake data. Simultaneously, there is evident similarity among the outcomes of the three image editing techniques employed.
Particularly, there is a higher population of edited images in correspondence of lower entropy values, while saliency maps of genuine samples exhibit higher entropy. 

Our findings bring to light an interesting outcome. While looking at altered images, humans tend to explore less when compared to genuine samples. 
We attribute this behavior to the inclination of individuals to focus more on specific details when encountering unfamiliar content within the image, in order to enhance their comprehension of the surrounding context. If we consider the proposed task, observers tend to rapidly shift their gaze from one location to another if they do not perceive anything unfamiliar, thus leading to higher degrees of disorders of the saliency maps. Sample saliency maps for both original and corresponding edited images are shown in Fig.~\ref{fig:teaser}.

\tit{Statistical Analysis} Given the real and fake entropy distributions, we further evaluate the results from a quantitative perspective. Specifically, we conduct the two-sample Kolmogorov-Smirnov (K-S) test in order to reveal a possible statistical difference between what is observed with the original images versus all the possible edits. Indeed, this is a non-parametric statistical test that evaluates whether the two sets of data come from the same population.
A larger statistic value indicates greater dissimilarity between the distributions. The K-S test is sensitive to differences in the tails of the distributions, making it a good choice to detect even minor discrepancies across the entire range.
We corroborate the outcomes of the K-S test by employing the two-sample Cramèr-von Mises (C-M) test. This test, akin to the K-S test, quantifies the divergence between the cumulative distribution functions (CDFs) of the two datasets. However, it does not solely concentrate on the maximum difference but instead considers the entirety of the CDF. It computes a test statistic that measures the overall discordance between the distributions, assigning more weight to disparities in the middle of the distributions, rendering it suitable for detecting variances in the central portion of the data.

Finally, evidence of the validity of our findings is obtained through the Mann-Whitney U (MWU) test, to assess the null hypothesis that two samples have the same central tendency.
In other words, this test is well-suited for comparing two independent samples when our aim is to establish whether one group typically exhibits greater values than the other. 

\begin{table}[t]
\caption{Statistical tests' results to evaluate the difference between entropy distributions across categories.
K-S, C-M and MWU refer to Kolgomorov-Smirnov, Cramér-von Mises and Mann Whitney U tests, respectively. Bolded results refer to significant $p$-values ($\alpha = 0.05$) and higher statistics.}
\label{tab:statistical_tests}
\vspace{-.15cm}
\setlength{\tabcolsep}{.32em}
\centering
\resizebox{\linewidth}{!}{
\begin{tabular}{cc c cc cc cc cc cc}
\toprule
& & & \multicolumn{2}{c}{\bf K-S Test} & & & \multicolumn{2}{c}{\bf C-M Test} & & &\multicolumn{2}{c}{\bf MWU Test}\\
\cmidrule{4-5} \cmidrule{8-9} \cmidrule{12-13}
\multicolumn{2}{c}{\bf Category}& & Statistic & $p$ & & & Statistic & $p$  & & & Statistic & $p$\\
\midrule
\multirow{3}*{O} & SA & & \bf{0.154} & $<$\bf{.001} & & & \bf{2.068} & $<$\bf{.001} & & & \bf{144409} & $<$\bf{.001}\\ 
 & SW & & \bf{0.106} & \bf{.007} & & & \bf{0.759} & \bf{.009} & & & \bf{136013} & \bf{.016}\\ 
 & IG & & \bf{0.166} & $<$\bf{.001} & & & \bf{2.807} & $<$\bf{.001} & & & \bf{148184} & $<$\bf{.001} \\ 
\midrule
 \multirow{2}*{SA} & SW & & 0.074 & .129 & & & 0.387 & .078 & & & 116543 & .064\\ 
 & IG & & 0.054 & .460 & & & 0.127 & .468 & & & 127714 & .552\\
\midrule
 SW& IG & & 0.086 & .050 & & & 0.690 & .013 & & & 136541 & .011\\
\bottomrule
\end{tabular}
}
\vspace{-.2cm}
\end{table}

Table~\ref{tab:statistical_tests} presents all the statistical results, including both the test statistic and its associated $p$-value.
In our analysis, if a $p$-value obtained from a test falls below the significance value $\alpha=0.05$, it indicates that the observed differences between the two datasets are statistically significant. More specifically, there is strong evidence to reject the null hypothesis and conclude that the two sets of data are different in a meaningful way. On the other hand, if the $p$-value is greater than $\alpha$, it suggests that the observed differences are not statistically significant, and we do not have sufficient evidence to reject the null hypothesis.

Comparing the entropy distribution of original images against semantic-agnostic, semantic-aware, and instruction-guided editing, the $p$-value is below the threshold in all cases and for all tests. As a consequence, we reject the null hypothesis that there is no significant difference between the distributions, highlighting that, in terms of entropy, there exists a distinguishable pattern in the perception of real and fake stimuli. By considering the difference between semantic-aware, semantic-agnostic, and instruction-guided editing classes, the null hypothesis is confirmed, meaning that these distributions can be assimilated into the same population of counterfeit images. The only exception stands between the semantic-aware and instruction-guided classes, where for both C-M and MWU tests a statistical difference exists. We argue that the primary reason lies in the output generation quality. As previously discussed, instruction-guided editing is the most easily detectable category as fake, while the opposite holds for the semantic-aware class. However, we are mainly interested in distinguishing between the real and all editing classes. In this case, the statistical difference between original and semantic-aware or between original and instruction-guided still holds.

\section{Conclusion}
Our exploratory study aimed to investigate the presence of an underlying pattern governing human visual perception when individuals view partially forged images in comparison to authentic ones. To facilitate this analysis, a novel dataset containing real images alongside their fake counterparts, both equipped with human eye fixations and ratings, was introduced.
Our findings reveal that when humans examine counterfeit images, their attention tends to be directed toward more confined regions, in contrast to genuine images where the observed visual pattern is more evenly distributed across the presented stimuli. Such results were supported through statistical tests conducted on the entropy distributions of the saliency maps, thereby confirming our initial hypothesis. We believe that our study could serve as a starting point for further research in the direction of semantics-based fake detection methods and, more in general, in the realm of human gaze-assisted artificial intelligence.

\bibliographystyle{IEEEtran}
\bibliography{bibliography}
\end{document}